\documentclass[10pt,twocolumn,letterpaper]{article}

\usepackage[pagenumbers]{cvpr} 

\definecolor{cvprblue}{rgb}{0.21,0.49,0.74}
\usepackage[pagebackref,breaklinks,colorlinks,allcolors=cvprblue]{hyperref}

\usepackage{cuted}
\usepackage[most]{tcolorbox}

\def\paperID{*****} 
\def\confName{CVPR}
\def\confYear{2026}

\title{Evaluating Reasoning Fidelity in Visual Text Generation}

\author{Jiajun Hong\\
Stony Brook University\\
{\tt\small jiajun.hong@stonybrook.edu}
\and
Jiawei Zhou \\
Stony Brook University\\
{\tt\small jiawei.zhou.1@stonybrook.edu}
}

\begin{document}
\maketitle
\begin{abstract}
Recent text-to-image (T2I) models can render highly legible and well-structured text within images, enabling applications including document generation and slide generation. However, it remains unclear whether such systems faithfully preserve reasoning ability when complex solutions must be expressed directly through rendered text, or whether they merely imitate surface-level patterns. We investigate this question by evaluating reasoning fidelity in visual text generation, where models must express complete reasoning processes as images. Our evaluation includes long text rendering, factual knowledge probing, context understanding, and multi-step reasoning. Across these settings, we find that current T2I models frequently produce semantic errors, logical inconsistencies, and incorrect intermediate steps, even when the rendered text appears visually clear. These failures contrast with the strong reasoning performance of text-only models on the same tasks. Our findings reveal a substantial gap between visual text generation and procedural reasoning, motivating more reliable visual text reasoning.

\end{abstract}
    
\section{Introduction}
\label{sec:intro}

\begin{figure}[t]
    \centering
    \includegraphics[width=\linewidth]{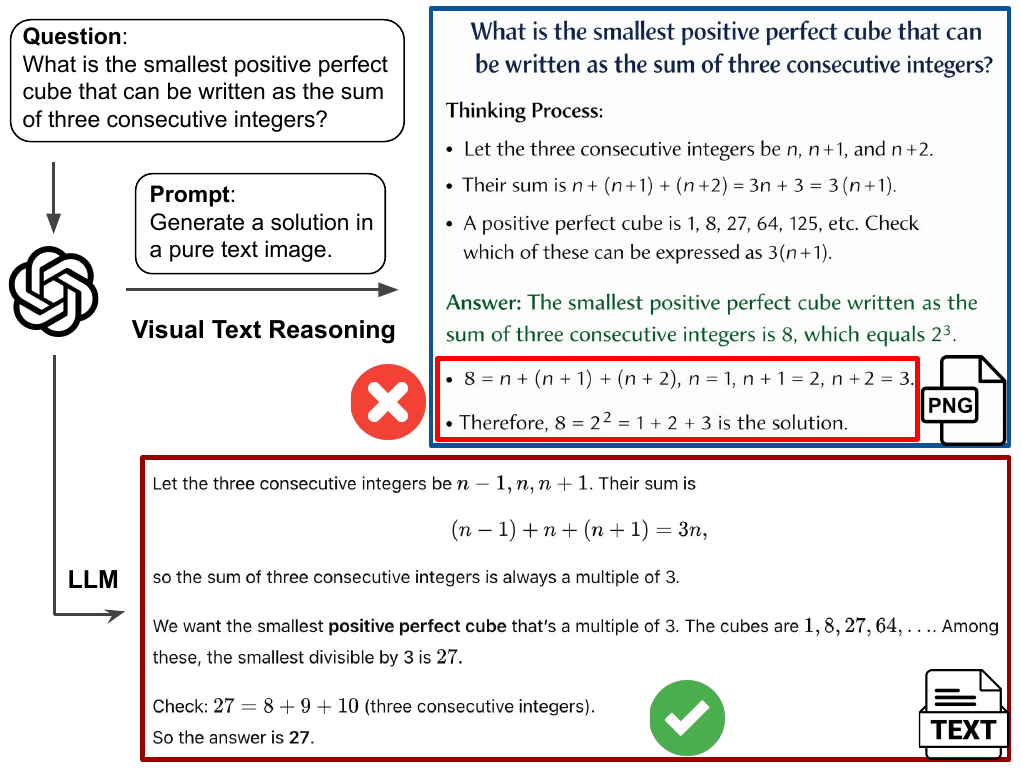}
    \caption{An example showing that, given the same reasoning problem, a T2I model (top) struggles to produce a correct solution, while LLM (bottom) can easily solve the task. Both outputs are generated using GPT-5.2.}
    \label{fig:teaser}
\end{figure}

Visual text generation has become an important capability of modern text-to-image (T2I) models, enabling applications such as document creation \citep{He2023DiffusionbasedDL}, slide generation \citep{zheng2025pptagent}, and interface design \citep{Wei2023BoostingGP, Wang2025DesignDiffusionHT} that require rendering textual content within images. Early diffusion-based models \citep{Ho2020DenoisingDP}, including SD-1.5 \citep{Rombach2021HighResolutionIS} and SD-XL \citep{Podell2023SDXLIL}, struggle to follow complex instructions and generate accurate text. Recent advances have substantially improved the ability of T2I models to produce legible visual text \citep{Tuo2023AnyTextMV,Chen2023TextDiffuserDM}. Existing evaluations primarily focus on rendering quality, measuring OCR accuracy, layout fidelity, and multilingual performance \citep{wang2025textatlas5m,zhao2025lex,yang2023glyphcontrol,wang2025uniglyph,liu2024glyph}. These studies show that performance degrades as text length increases \citep{zhang2025strict}. 
However, current evaluations mainly focus on text rendering quality rather than the semantic correctness.

Meanwhile, large language models (LLMs) have demonstrated remarkable capabilities in factual knowledge and multi-step reasoning 
\citep{Wei2022ChainOT,Lewkowycz2022SolvingQR, li2025context, DeepSeekAI2025DeepSeekR1IR, li2025okbench, feng-etal-2026-tracking, hossain2026hintmr}. As multimodal systems increasingly generate text-rich visual outputs, applications such as document generation and interface agents may require end-to-end visual outputs without accessible intermediate text. Therefore, visual quality alone is insufficient, and generated content must also remain semantically correct and logically consistent. In such settings, an important question arises: when multi-step solutions must be externalized through visual text, do T2I models retain the same semantic understanding and logical consistency observed in text-only settings?

We investigate this question through a diagnostic evaluation of reasoning fidelity under modality constraints. We design tasks that are straightforward for modern LLMs but challenging for T2I models when multi-step textual reasoning must be expressed visually, spanning factual knowledge recall and reasoning \citep{Clark2018ThinkYH}, long-context understanding \citep{Dua2019DROPAR}, and mathematical reasoning \citep{Hendrycks2021MeasuringMP}.
This evaluation is increasingly relevant as recent work has begun to blur the boundary between language and vision modalities.
For example, vision-language models (VLMs) have been shown to effectively process large amounts of textual information rendered as images, enabling text compression and long-context reasoning through visual inputs \citep{li-etal-2025-text, Wei2025DeepSeekOCRCO}. Conversely, recent T2I models are becoming increasingly capable of generating dense and structured textual content \citep{gptimage2}. Together, these developments suggest an emerging paradigm in which textual information can be represented, transmitted, and processed through visual channels, making it important to understand whether semantic reasoning remains faithful when expressed in visual form.


Because clear text rendering is a prerequisite for evaluating reasoning in visual text, we first introduce a preliminary literal long-text rendering task to identify models with sufficient visual text generation capability. We then disentangle rendering failures from reasoning failures through a layered evaluation protocol that separately measures text rendering errors, final-answer errors, and intermediate reasoning-step errors, supplemented with human validation. Across a variety of frontier and open-source T2I models, our experiments show that current models remain unreliable at producing logically coherent visual text, exposing a substantial gap between visual text generation quality and reasoning fidelity.

\section{Related Work}
\label{sec:related_work}

\paragraph{Text-to-Image Models for Visual Text Generation}
Accurate visual text generation remains a challenging problem for T2I models. Early diffusion-based approaches often struggle to render even short textual content correctly. To address this limitation, methods such as GlyphControl \citep{yang2023glyphcontrol} and AnyText \citep{Tuo2023AnyTextMV, Tuo2024AnyText2VT} incorporate glyph information and spatial constraints to improve layout controllability, while character-aware models \citep{liu2023character} and TextDiffuser \citep{Chen2023TextDiffuserDM, Chen2023TextDiffuser2UT} adopt character-level modeling to enhance text rendering quality. More recent approaches, including Glyph-ByT5 \citep{Liu2024GlyphByT5AC} and UniGlyph \citep{wang2025uniglyph}, further improve the generation of longer and more structured visual text. Despite these advances, existing methods primarily focus on rendering fidelity, layout control, and text accuracy, while largely overlooking whether generated visual text remains semantically correct and logically coherent.

\paragraph{Visual Text Generation Methods}
Several benchmarks have been proposed to evaluate visual text generation from different perspectives, including scene text synthesis, layout fidelity, multilingual rendering, and long-form text generation. Visual Text Generation in the Wild \citep{zhu2024visual} focuses on real-world scene text synthesis, while GlyphMM-3M \citep{wang2025uniglyph} evaluates character shape fidelity and layout consistency. LeX-Art \citep{zhao2025lex} emphasizes high-resolution text image synthesis and measures rendering quality and text attributes. Other datasets, such as TextAtlas5M \citep{wang2025textatlas5m} and STRICT \citep{zhang2025strict}, investigate the challenges of rendering long and dense text, showing that performance degrades as text length increases. Multilingual visual text generation has also been studied in \citep{liu2024glyph, zhang2025strict}. Collectively, these benchmarks have advanced the evaluation of visual text rendering; however, they primarily assess visual quality and text accuracy, leaving the semantic correctness and reasoning fidelity of generated visual text largely unexplored.

\paragraph{Visual Reasoning and Reasoning-Augmented Generation}
Some existing multimodal reasoning benchmarks, such as MMMU \citep{Yue2023MMMUAM}, MathVista \citep{Lu2023MathVistaEM}, and ChartQA \citep{Masry2022ChartQAAB}, evaluate a model's ability to reason over visual inputs, including charts, diagrams, and natural images. However, existing benchmarks and vision-language evaluation focus on interpreting visual content \citep{li2024socialgpt, chen2024halc, sadana2025iso, sheta2025behavioral, fang2026enhancing, kogilathota2026halp} rather than generating reasoning expressed through visual text.

Recent work has also explored enhancing reasoning in T2I models through external planning or reasoning modules. For example, ThinkDiff \citep{Mi2025ITT} aligns diffusion models with VLMs to improve multimodal reasoning, while GoT \citep{Fang2025GoTUR} and ShortCoTI \citep{gu2025improving} introduces chain-of-thought guidance and improves reasoning efficiency for image generation. Pipeline-based approaches such as PPAD \citep{Lv2025MultimodalLS} further decompose reasoning and generation into separate stages. Furthermore, novel usage of multimodal reasoning also explores processing textual information with vision-language models for improved accuracy and efficiency \citep{li-etal-2025-text, Wei2025DeepSeekOCRCO}. In contrast, our work studies a different question: whether T2I models can faithfully preserve semantic understanding and logical consistency when reasoning must be expressed directly through generated visual text.

\section{Task Design}
\label{sec:method}

Our goal is to evaluate the reasoning capability of T2I models when reasoning must be expressed through visual text. A key challenge is that failures in reasoning tasks may arise from incorrect reasoning or text rendering errors. To disentangle these factors, we first assess the text rendering fidelity of candidate models through a preliminary rendering task. Given a prompt $p$, a T2I model $M$ generates an image $I=M(p)$, from which we extract text $t=E(I)$. Models that fail to produce clear text are excluded from further reasoning evaluations.

\subsection{Preliminary Task: Text Rendering}
In this task, the model is asked to reproduce the input text exactly in image form ($t=p$). We randomly sample 500 passages from WikiText \citep{merity2016pointer},\footnote{\url{https://huggingface.co/datasets/Salesforce/wikitext}} a collection of verified and featured Wikipedia articles. 

To control task difficulty, we truncate each passage to a fixed word length of 64, 128, 256, and 512. To minimize style and background effects and ensure reliable text extraction, we use a standardized prompt template to generate a document-style image with plain text on a white background and no components in LaTeX-typeset style (Appendix~\ref{sec:appendix_prompts}).


\subsection{Advanced Reasoning Tasks}
Advanced reasoning tasks comprise long context understanding, factual knowledge and math reasoning in image space, representing progressively stronger reasoning requirements. These tasks evaluate whether T2I models with decent rendering ability can correctly perform multi-step reasoning. The prompt $p$ consists of context $c$, a question $q$, and an instruction $i$, $p=(c,q,i)$. The extracted text $t$ is expected to contain intermediate reasoning steps $r$ and final answer $a$, $t=(r,a)$. Detailed instructions can be found in Appendix~\ref{sec:appendix_prompts}.

\paragraph{Factual Knowledge}
This task evaluates whether the model exhibits basic factual knowledge. Prompts contain a question with four answer choices. The model is required to provide reasoning steps for each option and provide the final answer. We sample 200 examples each from the Easy and Challenge splits of the ARC dataset \citep{Clark2018ThinkYH}.\footnote{\url{https://huggingface.co/datasets/allenai/ai2_arc}} The questions consist of grade-school science problems that are typically straightforward for modern LLMs.

\paragraph{Context Understanding}
This task evaluates reasoning over long passages. Prompts consist of a long passage and an associated question. Models are required to derive the answer directly from the provided context while explicitly showing intermediate reasoning steps. We sample 300 data from the DROP dataset \citep{Dua2019DROPAR},\footnote{\url{https://huggingface.co/datasets/ucinlp/drop}} which requires discrete reasoning over the content of paragraphs, such as resolving references and performing simple operations like counting or addition.

\paragraph{Math Reasoning}
This task evaluates multi-step mathematical reasoning in visual space. Prompts contain a mathematical problem together with instructions to generate detailed reasoning and a final answer. We sample 500 data from the MATH dataset \citep{Hendrycks2021MeasuringMP},\footnote{\url{https://huggingface.co/datasets/DigitalLearningGmbH/MATH-lighteval}} with 100 for each difficulty level. The MATH dataset consists of competition-style math problems and is commonly used to evaluate and post-train large language models.



\section{Experiments}
\label{sec:experiments}

\subsection{Experimental Setup}
We evaluate state-of-the-art closed-source T2I models, including GPT-Image-1.5 \citep{gptimage15}, GPT-Image-2 \citep{gptimage2}, Gemini-2.5-Flash-Image \citep{gemini25flashimage}, and Flux.2-Pro \citep{flux2}. We also assess popular open-source models, including Qwen-Image \citep{Wu2025QwenImageTR}, SD-XL \citep{Podell2023SDXLIL}, and TextDiffuser-2 \citep{Chen2023TextDiffuser2UT}.
\citep{Podell2023SDXLIL},\footnote{\url{https://huggingface.co/stabilityai/stable-diffusion-xl-base-1.0}} and TextDiffuser-2 \citep{Chen2023TextDiffuser2UT}.\footnote{\url{https://github.com/microsoft/unilm/tree/master/textdiffuser-2}}

To compare different generation settings, we configure GPT-Image-1.5 with two generation qualities, low and medium, denoted as \textbf{GPT-L} and \textbf{GPT-M}. To avoid ambiguity, we abbreviate Gemini-2.5-Flash-Image, Qwen-Image, Flux.2-Pro, TextDiffuser-2, and SD-XL as \textbf{Gemini}, \textbf{Qwen-Img}, \textbf{Flux.2}, \textbf{TextDiffuser2}, and \textbf{SDXL}, respectively.

Since GPT, GPT-Image-2, Gemini, Qwen-Img, and Flux.2 are able to follow complex instructions, we design detailed prompts with explicit layout and formatting constraints (Appendix~\ref{sec:appendix_prompts}). In contrast, SDXL and TextDiffuser2 struggle with long instructions, so simplified prompts are used.

Images are generated using each model’s default resolution across all tasks: $1328 \times 1328$ for Qwen-Img, $512 \times 512$ for TextDiffuser2, and $1024 \times 1024$ for the remaining models. Larger resolutions do not provide additional benefits. We additionally include LLM baselines as a reference for reasoning performance without rendering constraints: the closed-source model GPT-5.2 and the open-source model Qwen3-8B.

\begin{table*}[t]
\centering
\small
\setlength{\tabcolsep}{4pt} 
\begin{tabular}{lccccccccc}
\toprule
\textbf{Model}
& \multicolumn{3}{c}{\textbf{Text Rendering}}
& \multicolumn{2}{c}{\textbf{Math Reasoning}}
& \multicolumn{2}{c}{\textbf{Context Understanding}}
& \multicolumn{2}{c}{\textbf{Factual Knowledge}} \\
\textbf{}
& CER$\downarrow$ & WER$\downarrow$ & ACC$\uparrow$ 
& $S_{a}\uparrow$ & $S_{p}\uparrow$ 
& $S_{a}\uparrow$ & $S_{p}\uparrow$ 
& $S_{a}\uparrow$ & $S_{p}\uparrow$ \\
\midrule
\textbf{GPT-5.2} & 0.0024 & 0.0024 & - & 0.934 & 0.969 & 0.870 & 0.936 & 0.988 & 0.994 \\
\textbf{Qwen3-8B} & 0.00004 & 0.0002 & - & 0.838 & 0.917 & 0.790 & 0.821 & 0.953 & 0.947 \\
\midrule
\textbf{GPT-Image-2} & 0.049 & 0.283 & 0.945 & 0.728 & 0.845 & 0.826 & 0.901 & 0.965 & 0.931 \\
\textbf{GPT-L} & 0.263 & 0.613 & 0.932 & 0.011 & 0.126 & 0.826 & 0.634 & 0.980 & 0.825 \\
\textbf{GPT-M} & 0.091 &0.347 & 0.963 & 0.520 & 0.615 & 0.803 & 0.822 & 0.945 & 0.919 \\
\textbf{Gemini} & 0.506 & 0.732 & 0.951 & 0.761 & 0.419 & 0.802 & 0.438 & 0.981 & 0.319 \\
\textbf{Qwen-Img} & 0.426 & 0.642 & 0.908 & 0.678 & 0.507 & 0.760 & 0.630 & 0.940 & 0.710 \\
\textbf{Flux.2} & 1.352 & 1.450 & 0.941 & 0.608 & 0.376 & 0.858 & 0.796 & 0.975 & 0.861 \\
\bottomrule
\end{tabular}
\caption{Overall performance of text-to-image models across all tasks. All results are averaged over the full evaluation set and across difficulty levels.}
\label{tab:overallresults}
\end{table*}

\begin{figure*}[t]
\centering
\includegraphics[width=0.8\textwidth]{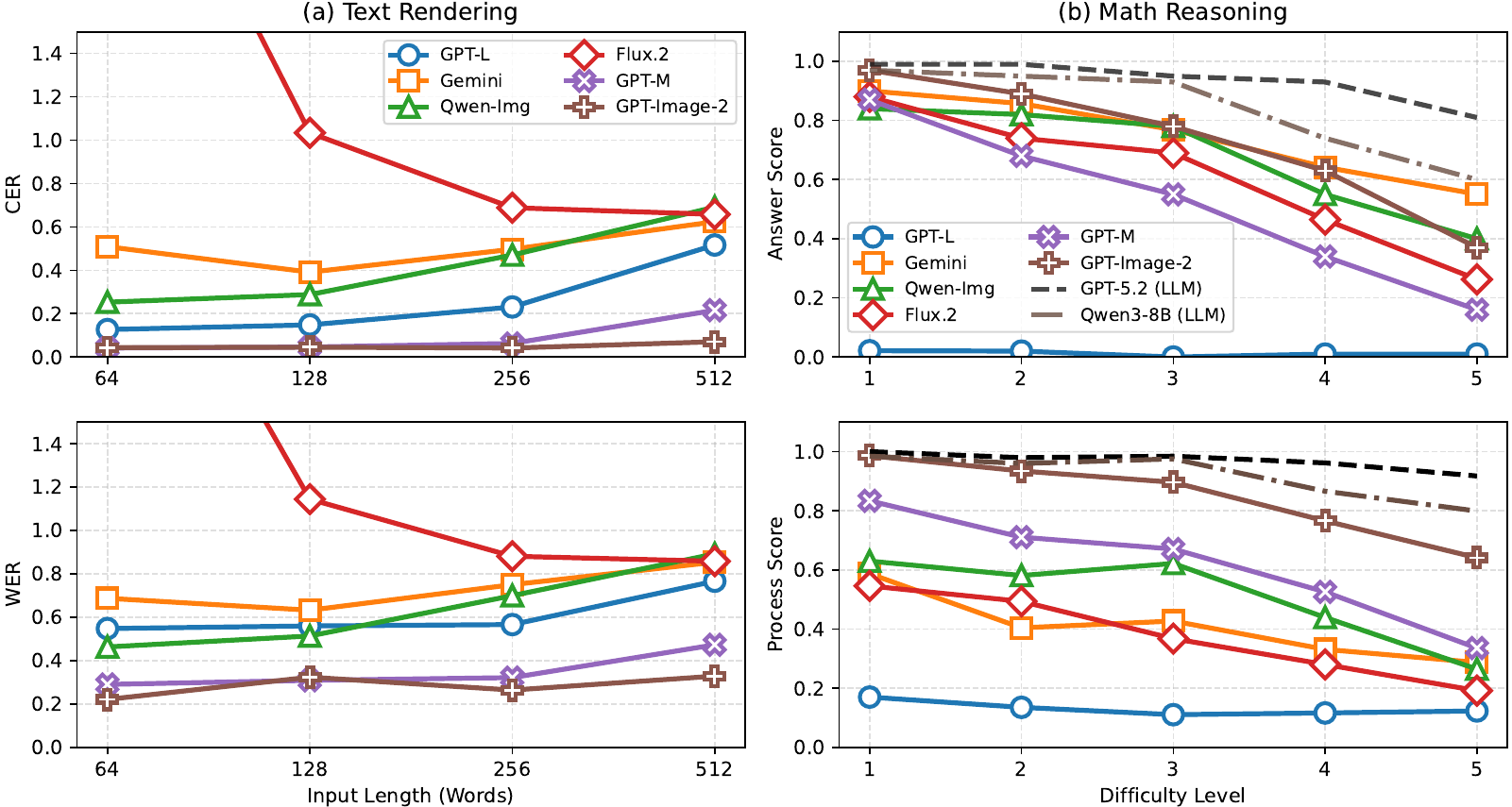}
\caption{
Complete performance visualization across tasks.
(a) Text rendering performance across increasing input lengths.
(b) Math reasoning performance across increasing difficulty.
}
\label{fig:results}
\end{figure*}

\subsection{Evaluation}
We evaluate models using a pipeline of prompt generation, image generation, text extraction with tools, and task-specific scoring.

\paragraph{Text Rendering}
We use PaddleOCR \citep{cui2025paddleocr30technicalreport}, which is commonly used in many other T2I evaluations works \citep{wang2025textatlas5m, zhao2025lex, yang2023glyphcontrol, wang2025uniglyph} to extract the text from the image and compare it against the original input text. Rendering fidelity is evaluated using Word Error Rate (WER), Character Error Rate (CER) \citep{zhang2025strict}, and OCR confidence (ACC), defined as the mean probability from OCR model to the predicted characters.

\paragraph{Advanced Reasoning}
Reasoning tasks are evaluated using an LLM-based extraction and assessment pipeline (GPT-5.2), since OCR models sometimes struggle to reliably extract mathematical expressions. Our reasoning tasks are straightforward for LLMs and extraction, which make LLM to be a reliable evaluator. Our ablation studies further show that extraction errors have limited impact on the final results.

Following the paradigm of process reward models \citep{Lightman2023LetsVS}, we evaluate reasoning at the step level. For mathematical reasoning, the evaluator receives the problem statement, all preceding reasoning steps, and the current step, and determines whether the current step is logically and mathematically valid. For factual knowledge and context understanding, the evaluator receives the context, question, and current reasoning step. Each step receives a binary score, and the process score $S_p$ is computed as the average across all steps. Because final answers may be corrupted by rendering errors or long reasoning traces, particularly in mathematical reasoning, we additionally evaluate answer correctness using an LLM judge and report an answer score $S_a$.

\section{Results and Analysis}
\label{sec:discussions}

\subsection{Text Rendering}

\begin{figure*}[t]
  \centering
  \includegraphics[width=\textwidth]{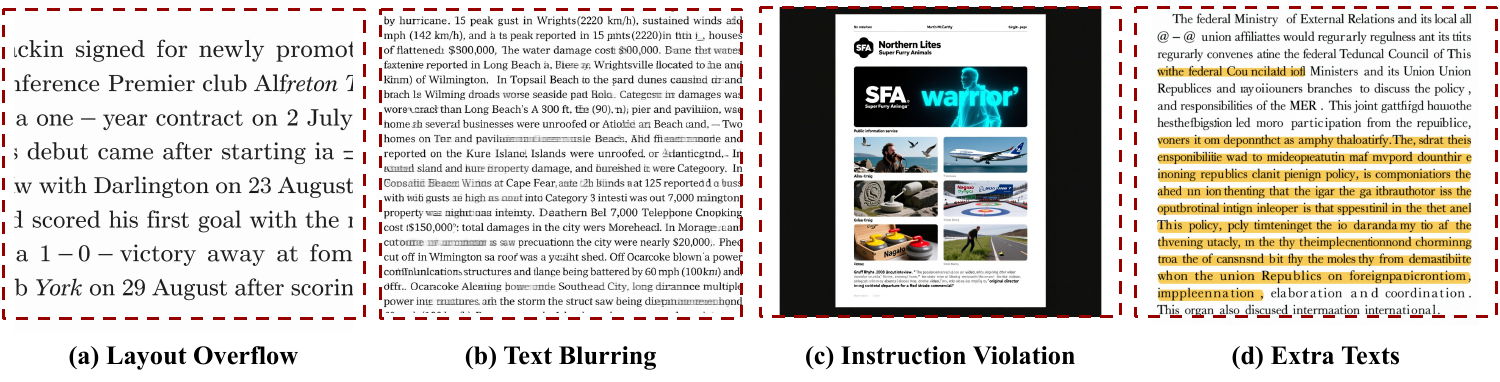}
  \caption{Illustration of common failure modes in the text rendering task. (a) An example generated by GPT-L. Poor layout planning causes characters near the image border to be cropped. (b) An example generated by GPT-L. Severe text blurring leads to low visibility. (c) An example generated by Qwen. The model fails to follow the instruction and generates unrelated content. (d) An example generated by Flux.2. The model extends the original text and hallucinates additional content. The highlighted text does not appear in the original text.}
  \label{fig:text_rendering_failure_flat}
\end{figure*}

\begin{figure*}[t]
  \centering
  \includegraphics[width=\textwidth]{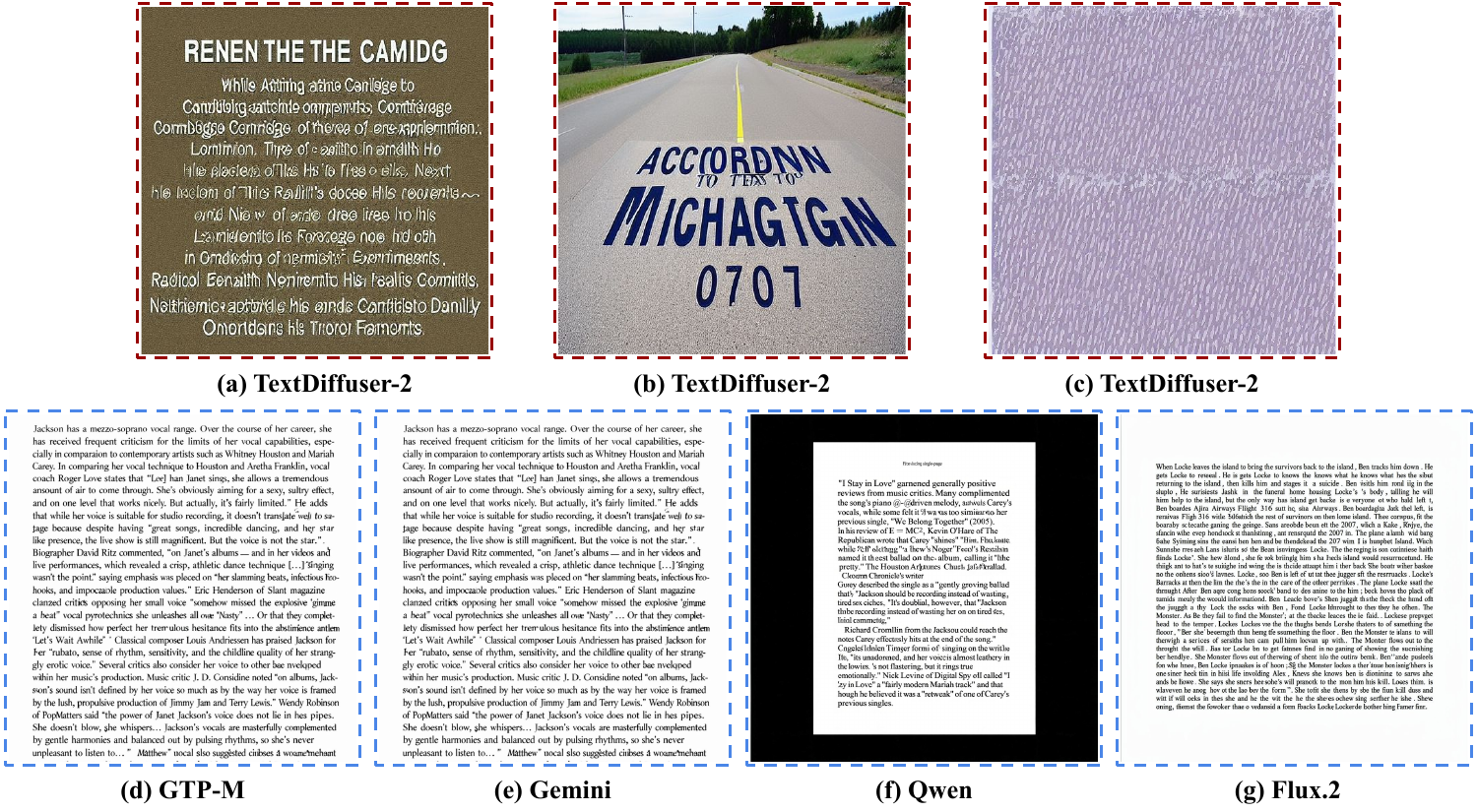}
  \caption{Examples of text rendering results from different models. (a) A failure case from TextDiffuser-2 where the rendered text is largely unrecognizable. (b) A failure case from TextDiffuser-2 where the model generates low-quality text fragments on an unrelated background image. (c) A failure case from TextDiffuser-2 for long input text, where the model fails to render the provided content. (d) A successful example from GPT-M with clear and correct text rendering. (e) A successful example from Gemini with clear and accurate text rendering. (f) A successful example from Qwen where the text is mostly clear with minor errors. (g) A successful example from Flux.2 where the text is clear with some errors.}
  \label{fig:text_rendering_cases}
\end{figure*}

Table~\ref{tab:overallresults} and Figure~\ref{fig:results} summarize the text rendering results. Despite recent advances in visual text generation, many T2I models exhibit systematic failures (Figure~\ref{fig:text_rendering_failure_flat}), including layout corruption, blurred text, instruction-following failures, and hallucinated content. These errors become increasingly severe as text length grows, suggesting that faithfully rendering long-form text remains challenging for current T2I systems.

Failure modes also vary substantially across models. GPT-M achieves relatively low CER but higher WER, suggesting that individual characters are often rendered correctly while word-level errors remain common. In contrast, models with high CER exhibit more severe character-level corruption. Flux.2 frequently generates additional hallucinated content beyond the provided text, while TextDiffuser2 and SDXL often fail even on short inputs (64 words), producing unrecognizable characters or unrelated images (Figure~\ref{fig:text_rendering_cases}). Due to their poor rendering fidelity, these models are excluded from subsequent reasoning evaluations.

\begin{figure*}[t]
  \centering
  \includegraphics[width=\textwidth]{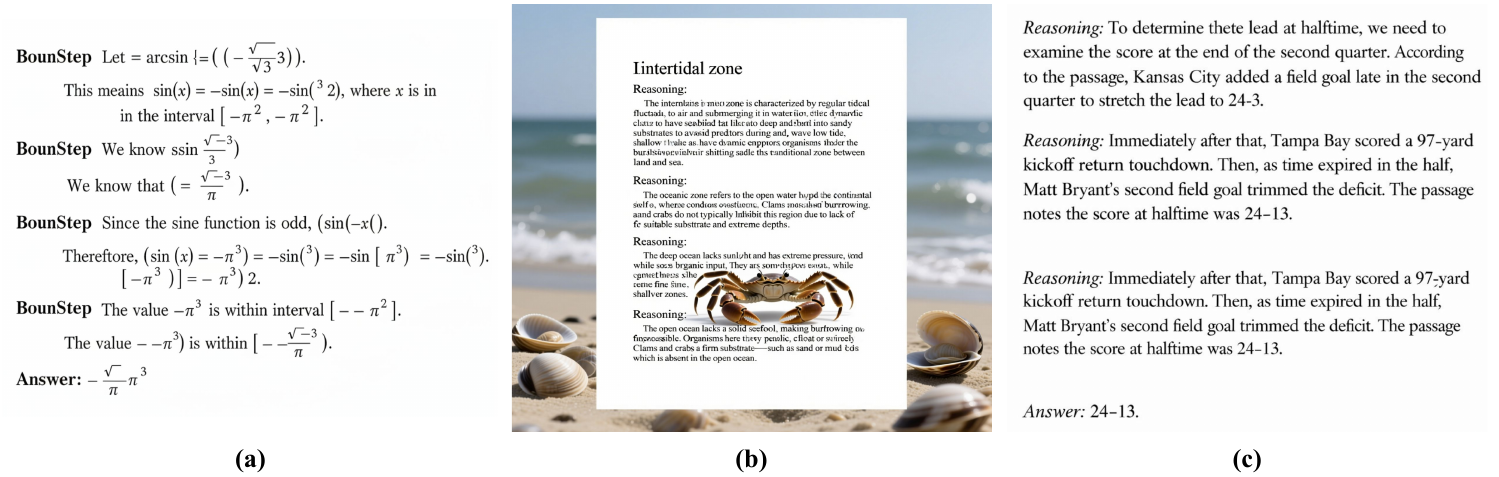}
  \caption{(a) An example generated by Gemini, illustrating irrational reasoning produced by a T2I model in math reasoning task. The generated reasoning steps are logically inconsistent and fail to support a valid answer. (b) An example generated by Qwen, hallucination in a reasoning task, where the generated image contains irrelevant content. (c) An example generated by Flux.2. A common failure case in reasoning tasks, where the model repeatedly reproduces previous steps, indicating an inability to perform coherent multi-step reasoning.}
  \label{fig:reasoning_failures}
\end{figure*}

\subsection{Advanced Reasoning in Visual Space}
\paragraph{Main Results}

Table~\ref{tab:overallresults} reports the main reasoning results. Among the evaluated T2I models, GPT-Image-2 consistently achieves the strongest performance across all tasks and exhibits the smallest gap between process scores and answer scores. This suggests that improvements in visual text generation can partially translate into stronger reasoning performance. However, there is still a noticeable gap relative to text-only LLM baselines, particularly for more difficult reasoning tasks.

Across all reasoning tasks, text-only LLMs significantly outperform T2I systems (Table~\ref{tab:overallresults}). GPT-5.2 achieves the highest performance overall, particularly in process scores, while the smaller Qwen3-8B also demonstrates strong reasoning consistency. Although some T2I models achieve relatively high answer accuracy, their process scores remain substantially lower, indicating that correct answers are often accompanied by flawed or inconsistent reasoning. Furthermore, performance consistently declines as task difficulty increases (Figure~\ref{fig:results}), with the largest gaps appearing on long-context understanding and mathematical reasoning tasks.

\begin{figure}[t]
    \centering
    \includegraphics[width=\linewidth]{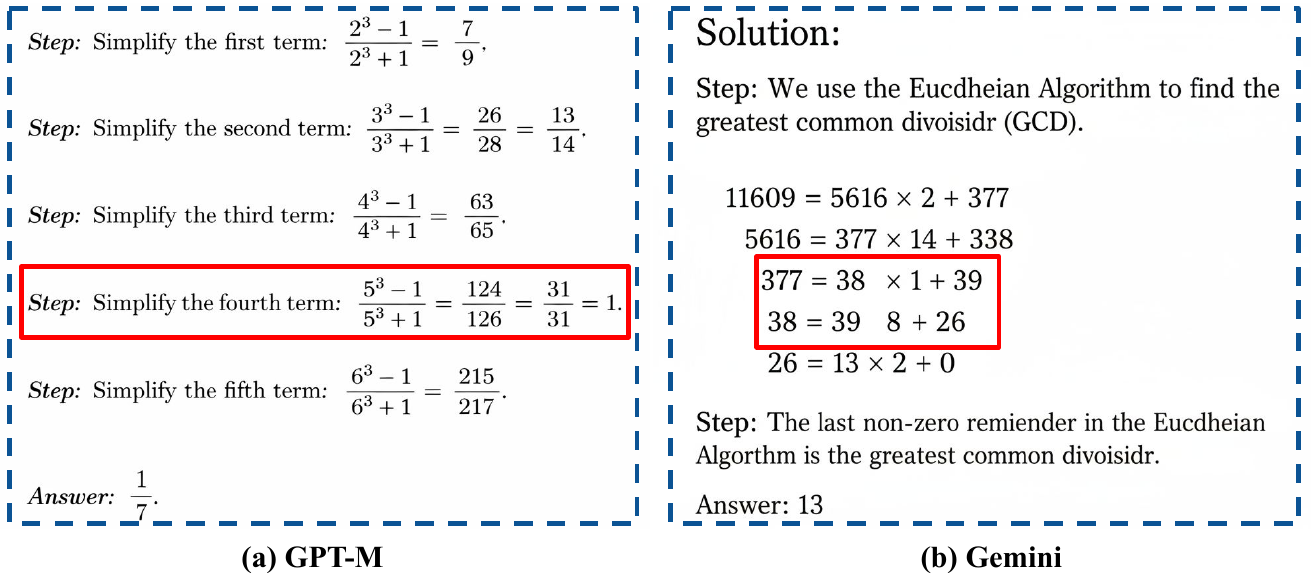}
    \caption{Examples where the rendered text is clear but reasoning steps (highlighted in red) are incorrect.}
    \label{fig:reasoning_failures_arr}
\end{figure}

Rendering issues such as unclear symbols or disordered layouts partially affect evaluation. However, reasoning failures are frequently observed even when characters are clearly rendered (Figures~\ref{fig:reasoning_failures_arr}, ~\ref{fig:reasoning_failures}), suggesting that the performance gap cannot be explained solely by rendering quality.

\paragraph{Performance Across Difficulty}
Performance consistently decreases as reasoning complexity increases (Figure~\ref{fig:results}). Although stronger T2I models such as GPT-Image-2 remain relatively robust, both process scores and answer scores decline substantially on more challenging tasks. For most other T2I models, the degradation in process scores is even more pronounced than the decline in answer scores. This gap suggests that correct answers do not necessarily imply correct reasoning. In many cases, models arrive at the correct solution despite producing flawed intermediate reasoning, revealing a persistent mismatch between answer correctness and reasoning fidelity.

\paragraph{Image Generation Quality Matters}
We further examine the relationship between generation quality and reasoning performance using GPT-L and GPT-M. GPT-M consistently outperforms GPT-L across all tasks, with the largest gains observed on reasoning-related metrics (Table~\ref{tab:overallresults}). Improved generation quality reduces rendering failures and leads to more coherent reasoning traces. In contrast, GPT-L frequently exhibits logical inconsistencies and incorrect intermediate steps. These results suggest that stronger visual text generation can partially improve reasoning performance, although it does not fully close the gap with text-only LLMs.

\paragraph{Large Reasoning Gap in Process}
Compared to the lightweight LLM Qwen3-8B, T2I models exhibit substantially lower process scores despite occasionally achieving comparable answer scores (Table~\ref{tab:overallresults}), indicating weaker reasoning consistency. The gap becomes even larger when compared with GPT-5.2, which significantly outperforms T2I models on both answer accuracy and reasoning. These results suggest that modern T2I models remain unreliable for rigorous multi-step reasoning, particularly when explicit intermediate steps are required.

\subsection{Error Source Analysis}
Overall process errors may arise from either rendering failures or reasoning failures. To better understand their relative contributions, we conduct ablation studies on OCR extraction and rendering quality.

\paragraph{OCR Backends}
We first examine whether OCR extraction significantly affects evaluation results. PaddleOCR is widely used for visual text generation benchmarks, but extraction errors could potentially confound rendering quality measurements. To verify robustness, we additionally evaluate the selected models using DeepSeek-OCR \citep{Wei2025DeepSeekOCRCO}.

As shown in Table~\ref{tab:ocr_backends}, DeepSeek-OCR produces consistent WER and CER results to PaddleOCR, indicating that OCR choice has limited impact on our evaluation. We therefore use PaddleOCR throughout the paper due to its faster inference speed and direct confidence estimates.

\begin{table}[t]
\centering
\begin{tabular}{lccccc}
\toprule
\textbf{Model}
& \multicolumn{3}{c}{\textbf{PaddleOCR}}
& \multicolumn{2}{c}{\textbf{DeepSeek-OCR}} \\

\textbf{}
& CER$\downarrow$ & WER$\downarrow$ & ACC$\uparrow$ 
& CER$\downarrow$ & WER$\downarrow$ \\

\midrule
\textbf{GPT-L} & 0.263 & 0.613 & 0.932 & 0.240 & 0.511 \\
\textbf{GPT-M} & 0.091 & 0.347 & 0.963 & 0.096 & 0.351 \\
\textbf{Gemini} & 0.506 & 0.732 & 0.951 & 0.502 & 0.716 \\
\textbf{Qwen} & 0.426 & 0.642 & 0.908 & 0.439 & 0.651 \\
\textbf{Flux.2} & 1.352 & 1.450 & 0.941 & 1.553 & 1.648\\
\bottomrule
\end{tabular}
\caption{Ablation study of OCR backends on the text rendering task.}
\label{tab:ocr_backends}
\end{table}

\paragraph{Rendering Errors}
We next investigate whether rendering quality is the primary source of low process scores. Using GPT-4.1, we estimate the readability of generated text through two metrics: Character Clear Rate (CCR), which measures the proportion of clearly readable characters, and All Clear Rate (ACR), which measures the proportion of images with fully readable text.

Results are reported in Table~\ref{tab:rendering_errors}. Most models achieve high CCR scores, indicating that the majority of generated characters are visually clear and recognizable. While some models, such as Qwen and GPT-L, exhibit lower ACR due to blurred characters or layout issues, models with high CCR and ACR scores still show substantial reasoning failures. Since individual reasoning steps are typically short, these results suggest that low process scores cannot be explained by rendering quality alone.

To validate the VLM-based metric, we manually annotate 200 samples from the text rendering task and 200 samples from the math reasoning task. The resulting CCR scores achieve a Pearson correlation of 0.785 with human judgments, increasing to 0.920 after removing 8 outliers (Appendix~\ref{sec:human_labels}). This result indicates that VLM-based CCR provides a reliable proxy for rendering quality in most cases.

\begin{table}[t]
\centering
\small
\begin{tabular}{lcccc}
\toprule
\textbf{Model}
& \multicolumn{2}{c}{\textbf{Text Rendering}}
& \multicolumn{2}{c}{\textbf{Math Reasoning}} \\

\textbf{}
& CCR$\uparrow$ & ACR$\uparrow$ 
& CCR$\uparrow$ & ACR$\uparrow$ \\

\midrule
\textbf{GPT-L} & 0.971 & 0.369 & 0.983 & 0.687 \\
\textbf{GPT-M} & 0.998 & 0.905 & 0.999 & 0.938 \\
\textbf{Gemini} & 0.988 & 0.758 & 0.996 & 0.899 \\
\textbf{Qwen} & 0.882 & 0.416 & 0.967 & 0.408 \\
\textbf{Flux.2} & 0.995 & 0.886 & 0.992 & 0.714 \\
\bottomrule
\end{tabular}
\caption{Ablation study of rendering errors.}
\label{tab:rendering_errors}
\end{table}

\subsection{Failure Cases}
\paragraph{Text Rendering Failure Cases}
Despite being explicitly instructed to reproduce the input text, T2I models frequently exhibit systematic rendering failures (Figure~\ref{fig:text_rendering_failure_flat}), including layout corruption, blurred text, instruction-following failures, and hallucinated content.

TextDiffuser2 and SDXL struggle even on short inputs due to limitations in visual text generation. With only 64 input words, they often produce unrecognizable characters, render only fragments of the provided text, or generate unrelated visual content. Performance further deteriorates as input length increases, leading to near-complete rendering failure on longer passages. In contrast, the models selected for subsequent reasoning experiments generally produce clear and readable text, as illustrated in Figure~\ref{fig:text_rendering_cases}.

\paragraph{Reasoning Failures}
Beyond rendering errors, we frequently observe reasoning failures even when the generated text is visually clear. Representative examples are shown in Figure~\ref{fig:reasoning_failures}. Common failure modes include logically inconsistent deductions, hallucinated intermediate steps, and repetitive reasoning patterns. Notably, these issues often occur even when the final answer is correct, indicating that answer correctness can mask substantial deficiencies in the underlying reasoning process.

\subsection{Limitations}
Our evaluation is intentionally end-to-end: a model must both generate an image and externalize its reasoning through rendered visual text. Although we mitigate extraction confounds via strict formatting constraints and validate robustness using alternative OCR backends and VLM-based readability checks, extraction errors—especially for mathematical symbols and equations—may still affect the scores. In addition, due to computational and budget constraints, we do not exhaustively sweep rendering-related factors such as font size, stroke width, or layout density, which may influence readability and extraction quality.

Another limitation is that many current text-to-image models are primarily optimized for generating natural images or short embedded text, rather than long-form structured documents. As a result, rendering dense paragraphs or multi-step reasoning traces remains challenging. Moreover, our evaluation focuses on reasoning expressed through explicit textual steps. Some models may internally perform reasoning but fail to faithfully externalize it in rendered text, which requires probing the internal states of these models.

In the future, it may be beneficial to develop specialized models or distilled variants of T2I systems that focus on reliable long-text rendering. Such models could provide a more stable platform for evaluating reasoning expressed through visual text and enable more reliable benchmarking of reasoning fidelity in the image space.

\section{Conclusion}
\label{sec:conclusions}

In this work, we investigate reasoning fidelity in visual text generation. We introduce a suite of evaluation tasks spanning long-text rendering, factual knowledge retrieval, long-context understanding, and multi-step mathematical reasoning, and evaluate whether modern T2I models can faithfully preserve reasoning when solutions must be expressed through rendered text. Our results show that, despite recent advances in visual text generation, current T2I models frequently produce logically inconsistent reasoning traces and incorrect intermediate steps, even when the rendered text is visually clear.

Overall, we identify a substantial gap between text rendering quality and reasoning fidelity. While improvements in visual text generation partially improve performance, current T2I models remain significantly less reliable than text-only LLMs at maintaining coherent reasoning processes. We hope this work motivates future research on reasoning-aware visual text generation and evaluation.

{
    \small
    \bibliographystyle{ieeenat_fullname}
    \bibliography{main}
}

\clearpage
\setcounter{page}{1}
\maketitlesupplementary

\newtcolorbox{promptbox}[1]{
  enhanced,
  breakable,
  width=2.0\linewidth,
  enlarge left by=\dimexpr 0.0\linewidth+0.5\columnsep\relax,
  colback=gray!10,
  colframe=gray!150,
  colbacktitle=gray!150,
  coltitle=white,
  fonttitle=\bfseries,
  title={#1},
  boxrule=1pt,
  arc=4pt,
  left=8pt,
  right=8pt,
  top=6pt,
  bottom=6pt
}

\section{Human Labels}
\label{sec:human_labels}
This section presents the details of how we manually label a subset of the data to verify that the VLM-based CCR metric is trustworthy in most cases.

For each selected model, we randomly sample 40 generated images across all difficulty levels in the text rendering task and 40 images across all difficulty levels in the math reasoning task. We manually annotate the clearness of characters and mathematical symbols in each image using a score between 0 and 1. A score of 0 indicates that no characters or symbols are clearly readable, while a score of 1 indicates that all characters and symbols are clearly rendered.

During annotation, we estimate the fraction of unclear characters relative to the total number of characters in the image. If a character is readable but slightly blurred, noisy, or not perfectly sharp, we count it as half a character rather than a fully clear one. This labeling process evaluates only the visual clearness of characters and symbols, without considering semantic correctness, logical consistency, or spelling errors.

\begin{figure}[t]
  \centering
  \includegraphics[width=0.9\columnwidth]{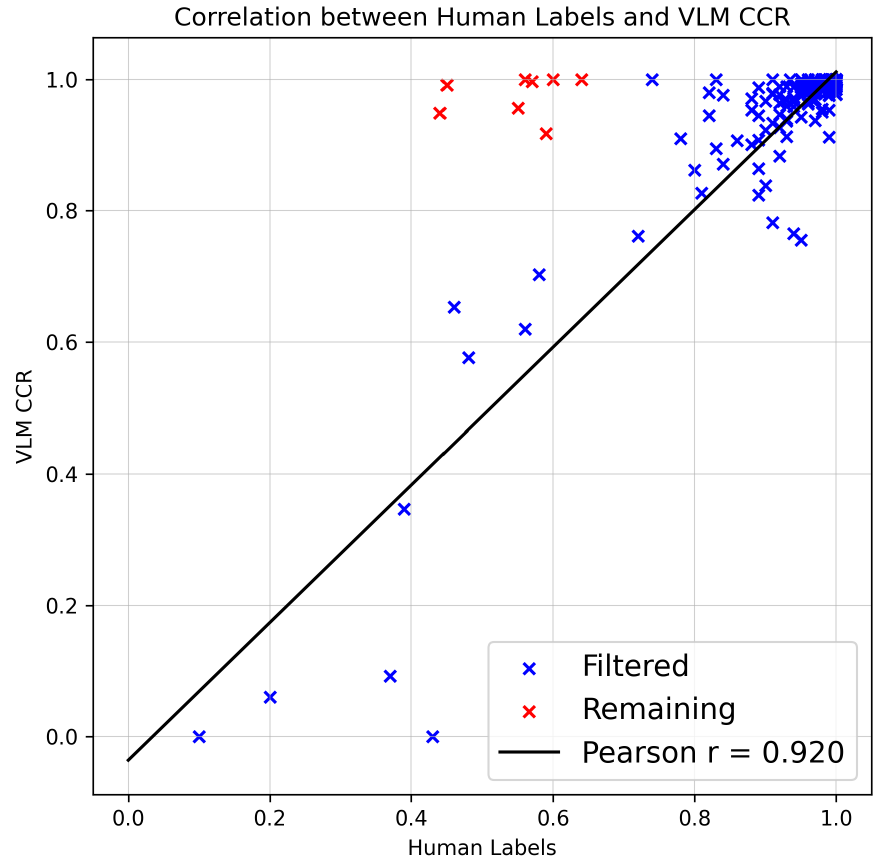}
  \caption{Pearson correlation between human annotations and the VLM-based CCR metric for character rendering quality.}
  \label{fig:human_label_correlation}
\end{figure}

We then compute the Pearson correlation between the VLM CCR results and our manually labeled scores. The Pearson correlation is 0.785 across all 400 samples. We observe that a small number of samples receive low human scores (more than 30\% characters unreadable) but high VLM CCR scores (about 0.95). We treat these as outliers. After removing these 8 samples, the Pearson correlation increases to 0.920 as shown in \cref{fig:human_label_correlation}. For all the 400 samples, the mean CCR is 0.972 and the mean label score is 0.954, which are very close. These result suggest that the VLM-based CCR metric provides a meaningful estimate of character rendering quality in most cases.

\section{Prompts}
\label{sec:appendix_prompts}

This section presents the detailed prompts used in our experiments. We first provide the prompts for image generation across our four tasks: Text Rendering, Context Reasoning, Factual Knowledge, and Math Reasoning in \cref{sub:image_generation_prompts}. We then provide the prompts used in evaluation, including those for scoring intermediate reasoning steps (process scores) and final answers in Factual Knowledge, Context Reasoning, and Math Reasoning tasks in \cref{sub:evaluation_prompts}. In addition, we provide the prompts used for VLM to check the clearness of the rendered text in \cref{sub:VLM verification prompts}.

\clearpage
\subsection{Image Generation Prompts}
\label{sub:image_generation_prompts}
\vspace{1.0\baselineskip}

\begin{promptbox}{Text Rendering Prompt}
    \setlength{\parskip}{1.0em}
    \setlength{\parindent}{0pt}
        
    Generate exactly one image. The image must be a flat, front-facing, single-page document with pure white background.
    Render only plain paragraph text, styled like a LaTeX-typeset document.
        
    Layout constraints:\\
    - Single-column layout only \\
    - No multi-column or vertical splits \\
    - Text flows continuously from top to bottom
        
    Formatting constraints:\\
    - Paragraph text only \\
    - No titles, headings, lists, or equations \\
    - No indentation beyond normal paragraph spacing
        
    Do not include LaTeX source code or markup.
    Do not depict any real-world scene.
    The visible text must match the following text exactly.
        
    Text: \{\textcolor{blue}{input\_text}\}
\end{promptbox}

\vspace{1.0\baselineskip}

\begin{promptbox}{Context Reasoning Prompt}
    \setlength{\parskip}{1.0em}
    \setlength{\parindent}{0pt}
        
    You are answering a reading comprehension question using the given passage. Return exactly one image. The image must contain only the solution text. Render plain paragraph text only in a single-column layout with pure white background. Style the text to resemble a LaTeX-typeset document (compiled appearance, not source code). Do not depict any real-world scene.
        
    Solution requirements: \\
    - Provide the reasoning first and then the final answer. \\
    - Each reasoning step must be one paragraph starting with "Reasoning:". \\
    - The final paragraph is one line, must start with exactly: "Answer: ". \\
    - After "Answer: ", output only the answer text. \\
    - If the answer has one span: output exactly that span. \\
    - If the answer has multiple spans: separate them with ", " (comma + single space). \\
    - If the answer requires arithmetic, output only the final number. \\
    - Otherwise, output one or more text spans copied verbatim from the passage. 
        
    Passage: \{\textcolor{blue}{passage}\} \\
    Question: \{\textcolor{blue}{question}\}
\end{promptbox}

\clearpage
\vspace{1.0\baselineskip}

\begin{promptbox}{Factual Knowledge Prompt}
    \setlength{\parskip}{1.0em}
    \setlength{\parindent}{0pt}
        
    You are answering a multiple choice question. 
    Return exactly one image. The image must contain only the solution text. Render plain paragraph text only in a single-column layout with pure white background. Style the text to resemble a LaTeX-typeset document (compiled appearance, not source code). Do not depict any real-world scene.
        
    Solution requirements: \\
    - Output only the reasoning steps and the final answer. \\
    - First output the reasoning and then the answer. \\
    - Provide exactly one reasoning paragraph for each choice. \\
    - Each reasoning step must be one paragraph starting with "Reasoning:". \\
    - The final paragraph is one line, must start with exactly: "Answer: <choice>". \\
    - The answer must be exactly only one capital letter: A, B, C, or D. \\
    - Do not include any other text after the answer line.
        
    Question: \{\textcolor{blue}{question}\}
\end{promptbox}

\vspace{1.0\baselineskip}

\begin{promptbox}{Math Reasoning Prompt}
    \setlength{\parskip}{1.0em}
    \setlength{\parindent}{0pt}
        
    Return an image contains only the solution text, typeset to look like a LaTeX-compiled math solution. Render only plain paragraph text, styled as a single-column layout LaTeX-typeset document with pure white background. Do not depict any real-world scene.
        
    The solution text inside the image must follow the exact structure:\\
    - Each reasoning paragraph starts with "Step:". \\
    - The final paragraph is the answer with only one line exactly: "Answer: <final answer>". \\
    - <final answer> contains ONLY the final result itself, with no other variables, no equations, no units, no description, and no explanatory text.
        
    Question: \{\textcolor{blue}{question}\}
\end{promptbox}

\clearpage
\subsection{Evaluation Prompts}
\label{sub:evaluation_prompts}

\vspace{1.0\baselineskip}

\begin{promptbox}{Factual Knowledge Process Score Prompt}
    \setlength{\parskip}{1.0em}
    \setlength{\parindent}{0pt}
        
    You are judging the correctness of one reasoning step, given the question and the reasoning step. Determine whether the current step is correct.
        
    Return only a single digit: \\
    - 1 if the current step is correct. \\
    - 0 if the current step is incorrect.
          
    Question: \{\textcolor{blue}{question}\} \\
    Reasoning steps: \{\textcolor{blue}{reasoning\_steps}\}
\end{promptbox}

\vspace{1.0\baselineskip}

\begin{promptbox}{Context Reasoning Process Score Prompt}
    \setlength{\parskip}{1.0em}
    \setlength{\parindent}{0pt}
        
    You are judging the correctness of one reasoning step, given the passage and the reasoning step. Determine whether the current step is correct given the passage.
        
    Return only a single digit: \\
    - 1 if the current step is correct. \\
    - 0 if the current step is incorrect.
          
    Passage: \{\textcolor{blue}{passage}\} \\
    Reasoning steps: \{\textcolor{blue}{reasoning\_steps}\}
\end{promptbox}

\vspace{1.0\baselineskip}

\begin{promptbox}{Context Reasoning Answer Score Prompt}
    \setlength{\parskip}{1.0em}
    \setlength{\parindent}{0pt}
        
    You are judging whether the candidate answer has the same meaning to the ground truth answer given the passage and the question.
        
    Return only a single digit and do not output anything else: \\
    - 1 if the given answer is equivalent to the ground truth. \\
    - 0 if the given answer is not equivalent.
        
    Passage: \{\textcolor{blue}{passage}\} \\
    Ground truth: \{\textcolor{blue}{ground\_truth}\} \\
    Candidate answer: \{\textcolor{blue}{candidate\_answer}\}
\end{promptbox}

\clearpage
\vspace{1.0\baselineskip}

\begin{promptbox}{Math Reasoning Process Score Prompt}
    \setlength{\parskip}{1.0em}
    \setlength{\parindent}{0pt}
        
    You are judging the correctness of one reasoning step, given the problem and all previous steps. Determine whether the current step is correct.
        
    Return only a single digit: \\
    - 1 if the current step is correct. \\
    - 0 if the current step is incorrect.
        
    Question: \{\textcolor{blue}{question}\} \\
    Previous steps: \{\textcolor{blue}{previous\_steps}\} \\
    Current step: \{\textcolor{blue}{current\_steps}\}
\end{promptbox}

\vspace{1.0\baselineskip}

\begin{promptbox}{Math Reasoning Answer Score Prompt}
    \setlength{\parskip}{1.0em}
    \setlength{\parindent}{0pt}
        
    You are judging whether the candidate answer is equivalent to the ground truth answer. Consider only the final answer content. Ignore formatting, wording, and explanation.
        
    Return only a single digit and do not output anything else: \\
    - 1 if the given answer is equivalent to the ground truth. \\
    - 0 if the given answer is not equivalent.
        
    Question: \{\textcolor{blue}{question}\} \\
    Ground truth: \{\textcolor{blue}{ground\_truth}\} \\
    Candidate answer: \{\textcolor{blue}{candidate\_answer}\}
\end{promptbox}

\clearpage
\vspace{1.0\baselineskip}
\subsection{Evaluation Prompts}
\label{sub:VLM verification prompts}
\begin{promptbox}{Verifying CCR with VLM}
    \setlength{\parskip}{1.0em}
    \setlength{\parindent}{0pt}
        
    You are evaluating text rendering quality in an image. Identify every visible character or symbol in the image, including: letters, numbers, punctuation and mathematical symbols. Count the total number of visible characters/symbols and how many of them are clearly and unambiguously readable by a human. Do not count whitespace.
        
    - Output exactly two integers separated by a single space. \\
    - First number: CLEAR \\
    - Second number: TOTAL \\
    - No words, no labels, no punctuation, no explanations. \\
    - Example valid output: 232 233
\end{promptbox}

\vspace{1.0\baselineskip}
\label{sub:VLM verification prompts}
\begin{promptbox}{Verifying ACR with VLM}
    \setlength{\parskip}{1.0em}
    \setlength{\parindent}{0pt}
        
    You are judging the text rendering quality from the given image. Are all the texts and symbols in the image rendered clearly and is readable by a human? Minor imperfections that do not affect reading are acceptable. Output exactly one character: 1 (yes) or 0 (no). Do not output anything else.

\end{promptbox}

\end{document}